\documentclass[a4paper]{article}

\usepackage{INTERSPEECH2016}

\usepackage{graphicx}
\usepackage{amssymb,amsmath,bm}
\usepackage{textcomp}
\usepackage{subcaption}
\usepackage{microtype}

\providecommand{\norm}[1]{\left\|#1\right\|}

\ninept

\usepackage{ifthen}

\usepackage[usenames,dvipsnames,dvinames]{xcolor}
\providecommand{\dodraft}{true}
\newboolean{isdraft}
\setboolean{isdraft}{\dodraft} 
\ifthenelse{\boolean{isdraft}}{%
\newcommand{\sunil}[2]{{\color{blue}{#1 $\to$ {\bf Sunil}}: #2}}
\newcommand{\jeff}[2]{{\color{red}{#1 $\to$ {\bf Jeff}}: #2}}
\newcommand{\todo}[1]{{\color{red}{{\bf TODO: #1} }}}
\newcommand{\fixme}[1]{{\color{red}{{\bf FIXME: #1} }}}

} {

\newcommand{\sunil}[2]{}
\newcommand{\jeff}[2]{}
\newcommand{\todo}[1]{}
\newcommand{\fixme}[1]{}

}

\title{Semi-Supervised Phone Classification using Deep Neural Networks and Stochastic Graph-Based Entropic Regularization}

\makeatletter
\def\name#1{\gdef\@name{#1\\}}
\makeatother \name{{\em Sunil Thulasidasan$^{1,2}$, Jeffrey Bilmes$^2$}}

\address{$^1$Los Alamos National Laboratory\\
  $^2$Department of Electrical Engineering, University of Washington \\
  {\small \tt sunil@lanl.gov, bilmes@uw.edu}
}

\begin{document}
  \maketitle

 \begin{abstract}
We describe a graph-based semi-supervised learning framework in the context of deep neural networks that uses a graph-based entropic regularizer to favor smooth solutions over a graph induced by the data. 
The main contribution of this work is  a computationally efficient, stochastic  graph-regularization technique that uses mini-batches that are consistent with the graph structure, but also provides enough stochasticity (in terms of mini-batch data diversity) for convergence of stochastic gradient descent methods to good solutions. For this work, we focus on results of frame-level phone classification accuracy on the TIMIT speech corpus  but our method is general and scalable to much larger data sets.  Our method significantly improves classification accuracy in the low-labeled scenario, and it is competitive with other methods in the fully labeled case. 

  \end{abstract}

\section{Introduction}
\label{sec:intro}
Semi-supervised learning (SSL) methods use  both labeled and unlabeled data to improve learning performance~\cite{chapelle2006semi}  and are especially useful in situations where labeled data is scarce. 
Since unlabeled data can usually be  collected in a fully automated, scalable way, SSL methods  aim to leverage unlabeled data to improve prediction performance by exploiting the similarity between labeled and unlabeled data.  A natural  way to capture this relationship is via graphs where the nodes represent both labeled and unlabeled points and the weights of the edges reflect the similarity between the nodes~\cite{zhu2005semi}. The main idea behind  graph-based SSL methods is that 
the objective function constrains the output of the classifier to be similar for nodes that lie close to each other on the manifold.
Graph-based SSL algorithms have been successfully applied to tasks such as  phone and word classification in automatic speech recognition (ASR)~\cite{malkin2009semi, liu2013graph,liu2014graph,liu2015acoustic},  part-of-speech  tagging~\cite{subramanya2010efficient}, statistical machine translation~\cite{alexandrescu2009graph}, sentiment analysis in social media~\cite{lerman2009sentiment}, text categorization~\cite{subramanya2008soft} and many others.  

In this work, we describe algorithmic improvements for efficient and scalable graph regularization that can be applied to any parametric graph-based SSL framework. We use  a fully parametric learner -- a deep neural network --  with an entropy regularizer over the graph induced by the data, a method that was first described in~\cite{malkin2009semi} in the context of a  multi-layered perceptron (MLP) with one hidden layer. By sampling the data using graph partitioning, but at the same time preserving the statistical properties of the data distribution,  and by stochastically regularizing over the graph,  we are able to significantly outperform the original results even on an MLP, and make further improvements using a DNN. For the results reported in this paper,  we limit our  data-set to fixed length speech frames, only reporting frame-level phone classification accuracy on the TIMIT\cite{garofolo1993darpa} speech corpus without using  HMM-based decoding and $n$-gram language models.  Our aim in this work is not to beat state-of-the-art ASR systems (which all use language models\cite{abdel2012applying, mohamed2009deep, palaz2013estimating} and typically will have higher accuracy than the results presented here) but to demonstrate the efficacy of a computationally efficient technique that can potentially be used  to improve ASR systems in a semi-supervised setting.

\section{Parametric Objective for Graph-based SSL Classifiers}
\label{sec:objective}

Graph-based SSL techniques assume that data are embedded in some
low-dimensional manifold in a higher dimensional ambient space, and
that nearby nodes will likely have the same labels (the manifold and
smoothness assumptions, respectively). Let 
$\{(\bf{x}_i,y_i)\}_{i=1}^{\ell}$ be the labeled training data
and 
$\{\bf{x_i}\}_{i=\ell+1}^{\ell + u}$ be the unlabeled training data,
where $n = \ell + u$ so that we have $n$ points in total.  We denote
by $\vartriangle_M$ the $M$-dimensional probability simplex (i.e., the
set of all distributions over $M$ class labels). Let
$\mathbf{p}_\theta (\mathbf{x}_i) \in \vartriangle_M$ represent the
output vector of posterior probabilities dictated by $\theta$, the
parameters of the classifier and $\mathbf t_i \in \vartriangle_M$ for
$1 \leq i \leq \ell$ denote a probabilistic label vector for the
$i$-th training sample.  We also assume that the samples $\{ \mathbf{x}_i\}_i$ are
used to produce a weighted undirected graph $\mathcal{G} = (V,E, \bf W)$, where
$\omega_{i,j} \in \bf{W}$ to be the similarity (edge weight)
between samples (vertices) $\mathbf x_i$ and $\mathbf x_j$ ($i$ and $j$).
We use the objective function defined 
in~\cite{subramanya2009entropic,malkin2009semi}, namely:
\begin{align}
J(\theta) &= \sum_{i=1}^l  \mathbf{D}( \mathbf{t}_i  \parallel \mathbf{p}_\theta (\mathbf{x}_i))  +
\gamma \sum_{i,j=1}^n \omega_{i,j} \mathbf{D}(\mathbf{p}_\theta (\mathbf{x}_i) \parallel \mathbf{p}_\theta (\mathbf{x}_j))   \nonumber \\
 &+ \kappa \sum_{i=1}^n \mathbf{D}(\mathbf{p}_\theta (\mathbf{x}_i) \parallel
\mathbf{u})  + \lambda \norm{\theta},   \label{eq:loss_fn}
\end{align}
where $\mathbf{u} \in \vartriangle_M$ is the uniform distribution and  $J(\theta)$  is the loss calculated over all samples. 
 The first term in the above equation is the supervised KL-divergence loss over the training samples, and the second term is the penalty imposed by the graph regularizer over neighboring pairs of nodes that favors smooth solutions over the graph.  The  third term  is an entropy regularizer and favors higher entropy distributions since MLPs and DNNs are often very confident in their predictions which can lead to degenerate solutions;  favoring higher entropy solutions counters this and is especially useful near decision boundaries. An alternative to regularizing against the uniform distribution is to regularize against a prior $ \bf{ \tilde{p}} ( \mathbf{x}_i )$ as done in~\cite{liu2013graph}, where $ \bf{ \tilde{p}} ( \mathbf{x}_i )$ are the outputs from a first-pass classifier trained in a supervised manner. We can easily incorporate this into our framework, though the work described in this report only  regularizes against the uniform distribution.  The final term in Equation~\ref{eq:loss_fn} is the standard  $\ell_2$ regularizer to discourage overfitting.

 \section{Related Work}
 \label{sec:related}
 
 There have been several graph-based learning algorithms that make use of some version of the objective function described in the previous section~\cite{liu2013graph, subramanya2009entropic,  zhu2002learning, talukdar2009new}.  Label propagation, described in~\cite{zhu2002learning} forces $f$ to agree with labeled instances by minimizing squared loss between predictions of nearby points. Measure propagation, described in~\cite{subramanya2009entropic}  uses essentially the same objective function as in Equation~\ref{eq:loss_fn} but in a non-parametric setting. Prior-regularized measure propagation~\cite{liu2013graph} substitutes the uniform distribution in Equation~\ref{eq:loss_fn} with a prior $\bf{\tilde{p}}_i$ that comes from a supervised classifier prior to the SSL process, and has shown to perform well on speech data.  One of the early works to use a graph regularizer in a deep learning context is described in ~\cite{weston2012deep}, where squared loss is used instead of KL-divergence. The algorithms described in this paper  will generally work on any objective function with a graph regularizer.
 

\section{Graph Regularization via Graph Partitioning}
Like other graph-based SSL methods we induce a graph on the data  by constructing a $k$-nearest neighbor ($k$-NN) graph  where the edge weights are the Euclidean distance between the feature vectors.  
 We use stochastic gradient descent (SGD) with mini-batches to optimize our objective function and use 
 larger mini-batches (size set to 1024) for better computational efficiency. Traditional SGD methods require randomly shuffling of the data for good convergence before constructing the mini-batches whic is problematic for our objective function. To see this,  consider the terms involving graph regularization from our objective function, calculated over each point:
 \begin{align}
G_i =  \gamma \sum_{j=1}^n \omega_{i,j} D(\mathbf{p}_\theta (\mathbf{x}_i) \parallel \mathbf{p}_\theta (\mathbf{x}_j))   \nonumber
\end{align}
For a randomly shuffled data-set, given that the $k$-NN graph is very sparse 
 the chunk of the affinity matrix corresponding to the mini-batch will be extremely sparse, implying that graph regularization will fail to take place on most computations.

For the graph regularizer to be effective in a computationally efficient way, our mini-batches need to reflect the structure of the graph. To do this, we partition our affinity graph into $k$ balanced parts by minimizing edge-cut (i.e, given $k$,  we want to minimize the number of edges between partitions). The resulting re-permuted affinity matrix has a dense block-diagonal structure; during mini-batch gradient descent, each mini-batch tends to correspond to  points inside one of the partition blocks. The corresponding relatively dense sub-matrices of the affinity matrix are used for the graph regularization computation over the mini-batches.


%

\subsection{SGD on Graph-Based Mini-Batches}
Theoretically, SGD, subject to certain assumptions, gives us an unbiased estimate of the true gradient. To see this, following the argument in~\cite{bengio2012practical}, consider the generalization error $C$ of some learner with parameters $w$, classification function $f_w$, and loss function $L$. The generalization error and gradient computed over a sample $(x,y)$ are respectively:
 \begin{align}
 &C = E[L(x,w)]  = \int p(x)L(f_w(x,y))dx \\
 &\hat g_w(x) =  \frac {\partial L(f_w(x,y))}{\partial w}
 \end{align}
 The gradient of the generalization error is 
 \begin{align}
 \frac{\partial C}{\partial w} &= \frac{\partial}{\partial w} \int p(x)L(f_w(x,y))dz  = \int p(x) \frac {\partial L(f_w(x,y))}{\partial w} dx\\
 &  = \int p(x)  \hat g_w(x) = E[\hat g_w]
 \end{align}
meaning $\hat g_w$ is an unbiased, albeit noisy, estimate of the gradient of the generalization error. 
We can smooth out the noise of the gradient by using larger mini-batches during gradient computation, where the gradient estimate in the mini-batch case for some mini-batch $S_k$ is given by
\begin{align}
\hat g_w(x) = \frac{1}{|S_k|} \sum_{i \in S_k}  \frac{\partial L(f_w(x_i,y_i))}{\partial w}
\end{align}
Using linearity of expectations, it is easy to show that the mini-batch gradient is also an unbiased estimate of the true gradient.
Note, however, that the above argument is based on the data being sampled from the true distribution $p(x)$. If our entire data set approximates this true distribution reasonably well, then a randomly sampled mini-batch will also be faithful to this distribution. However, for a graph partitioned mini-batch this argument no longer holds since the data points that comprise a mini-batch are  not randomly sampled, but on the contrary, reflect  relatively homogeneous regions (since we are partitioning by minimizing edge-cut on a $k$-NN graph) on some low dimensional manifold. Thus our gradient estimate is no longer unbiased, leading to poor convergence of SGD. On the other hand we have also seen that randomly shuffled batches will cause the graph regularizer to become ineffective due to poor within-batch neighbor connectivity, unless one accepts extremely long computational times (and communication costs in a parallel implementation).

Constructing a mini-batch that gives good SGD convergence, and good neighbor connectivity represents a trade-off between  two  somewhat mutually opposing properties: diversity (for SGD convergence, also found to be the
case in \cite{kaiwei2015nipsparallelworkshop,kaiwei2015nips_submod_partitioning}
) and good neighbor-connectivity (for efficient graph regularization), which usually implies homogeneity. The full batch (i.e., the entire data set) however, has both these properties; perfect neighbor connectivity (since it contains all the points) as well as diversity\footnote{We use Shannon entropy, calculated on the labels in a mini-batch, as a measure of diversity, but we anticipate better diversity measures exist.
\looseness-1
} that mimics the diversity within the complete training data (assuming a large enough, well sampled training set).  Indeed, if we are allowed to increase the size of the mini-batches as we please, we could presumably capture a more diverse set of points as well as  a significant fraction of their neighbors, but computational and memory constraints prevent us from doing so.  Note that the global structure of the affinity graph, owing to its sparsity, consists of a large number of small tightly connected clusters, with relatively few edges between the clusters. Thus a mini-batch that somehow captures this structure, but on a smaller scale, will  be expected to have reasonably good connectivity as well as high entropy. This suggests a possible heuristic recipe for the construction of improved mini-batches:
\begin{enumerate}
\item Given $N$ data points, a  batch size $B$ (that  represents our memory constraint) and $M$ classes, partition the entire graph into $\frac{N M}{B}$ mini-blocks, where each mini-block is approximately balanced at size $B/M$.
\item Construct $N/B$ ``meta" batches of  size $B$ from the mini-blocks as follows:
\begin{enumerate}
\item For each batch $b_i$, randomly choose $M$ mini-blocks from the set of $\frac{NM}{B}$  mini-partitions that were created in Step $1$. 
\item Group these $M$ mini-blocks into one larger meta-batch. Since each mini-block is approximately size $B/M$, our meta-batch will be approximately of size $B$, satisfying our memory constraint.
\end{enumerate}
\end{enumerate}
At the end of this process we have meta-batches which are  of the same size $B$ as the earlier graph-based batches, but which are qualitatively different. Each meta-batch is now composed of many small homogeneous mini-blocks which, due to random sampling,  are likely to be of a different class. We omit the proof here due to space constraints,  but intuitively we expect that the resulting entropy from grouping together $M$ such randomly chosen mini-blocks (of approximately equal size) to approach the entropy of the training set. 

To see the effect of this process on the within-batch neighbor connectivity of the meta-batch, let $\mathcal{N}_i$ represent the set of neighbors of node $i$ and $\mathcal{C}_i \subseteq \mathcal{N}_i$ be the set of neighbors of a node $i$ that are within the same batch. Let $\mathcal{M}_j$ be the set that represents mini-batch $j$.  We use the following definition for within-batch connectivity of $\mathcal{M}_j$:
\begin{align}
 c_j = \frac{\sum_{i \in \mathcal{M}_j} |\mathcal{C}_i   |}{\sum_{i \in \mathcal{M}_j} |\mathcal{N}_i|}, j = {1,2,3 \dots k}  \label{eq:connectivity}
\end{align}
Let $C_{mini}$ and $C_{meta}$ denote the random variables that represent the within-batch connectivity of a mini-block and meta-batch respectively. One can show that grouping $K$ mini-blocks to form a meta-batch does not adversely impact the connectivity score, i.e.,~$E[C_{meta}] \geq E[C_{mini}]$. Further, using the Central Limit Theorem, we can show that the variance of  $c_{meta}$ is given by $\sigma^2_{c_{meta}} =  \frac{1}{K} \sigma^2_{c_{mini}}$.

\subsection{Stochastic Regularization over Graphs}
Even though a meta-batch constructed using the procedure described in the previous section has much better neighbor-connectivity than a randomly shuffled batch, for a given node, a significant number of neighbors still lie outside the meta-batch.\footnote{This fraction will depend on the mini-block size; for most of the experiments in this paper about $30\%$ of the neighbors lie within a meta-batch.}
 As we argued earlier, regularizing against all neighbors is computationally inefficient. To preserve efficiency while still regularizing against out-of-batch neighbors, at each step, we {\em randomly} pick one additional meta-batch and regularize against this neighbor as follows:  consider the graph induced by the meta-batches, $\mathcal{G}_M = (V^M,E^M)$,  $V^M =  \left \{  M_1,M_2...M_{\lfloor N/B \rfloor } \right \}$ where each $M_i$ is a meta-batch, and edge $e_{i,j}^M \in E^M$ exists between $M_i$ and $M_j$ if there exist some edge $e_{s,t}$  between nodes $v_s$ and $v_t$  in the affinity graph $\mathcal{G}$, such that $v_s \in M_i$ and $v_t \in M_j$. That is, meta-batches are connected if their member nodes are connected in the original affinity graph. Let $\mathcal{C}_{i,j}$ denote the set consisting of all such unique pairs ${v_s,v_t}$. Then we can define an edge-weight on each of the edges in $E^M$  as $\vert  \mathcal{C}_{i,j} \vert $. For a given meta-batch $M_j$, during each epoch, the probability of picking a neighboring meta-batch $M_j$ is given by
\begin{align}
p_{i,j} = \frac{\vert  \mathcal{C}_{i,j} \vert} {\sum_{j} \vert  \mathcal{C}_{i,j} \vert}.
\end{align}
Thus, a neighboring batch $M_j$ of batch $M_i$ is more likely to be picked during an epoch if there are a relatively large number of edges between the member nodes that comprise $M_i$ and $M_j$. Over a large number of epochs, graph regularization is likely to take place against all neighboring batches; this enables labels to propagate via a stochastic diffusion process within the connected components of the affinity graph.

\section{Experiments}
\begin{figure*}
\centering
	\begin{subfigure}{0.66\columnwidth}
		\includegraphics[width=\columnwidth]{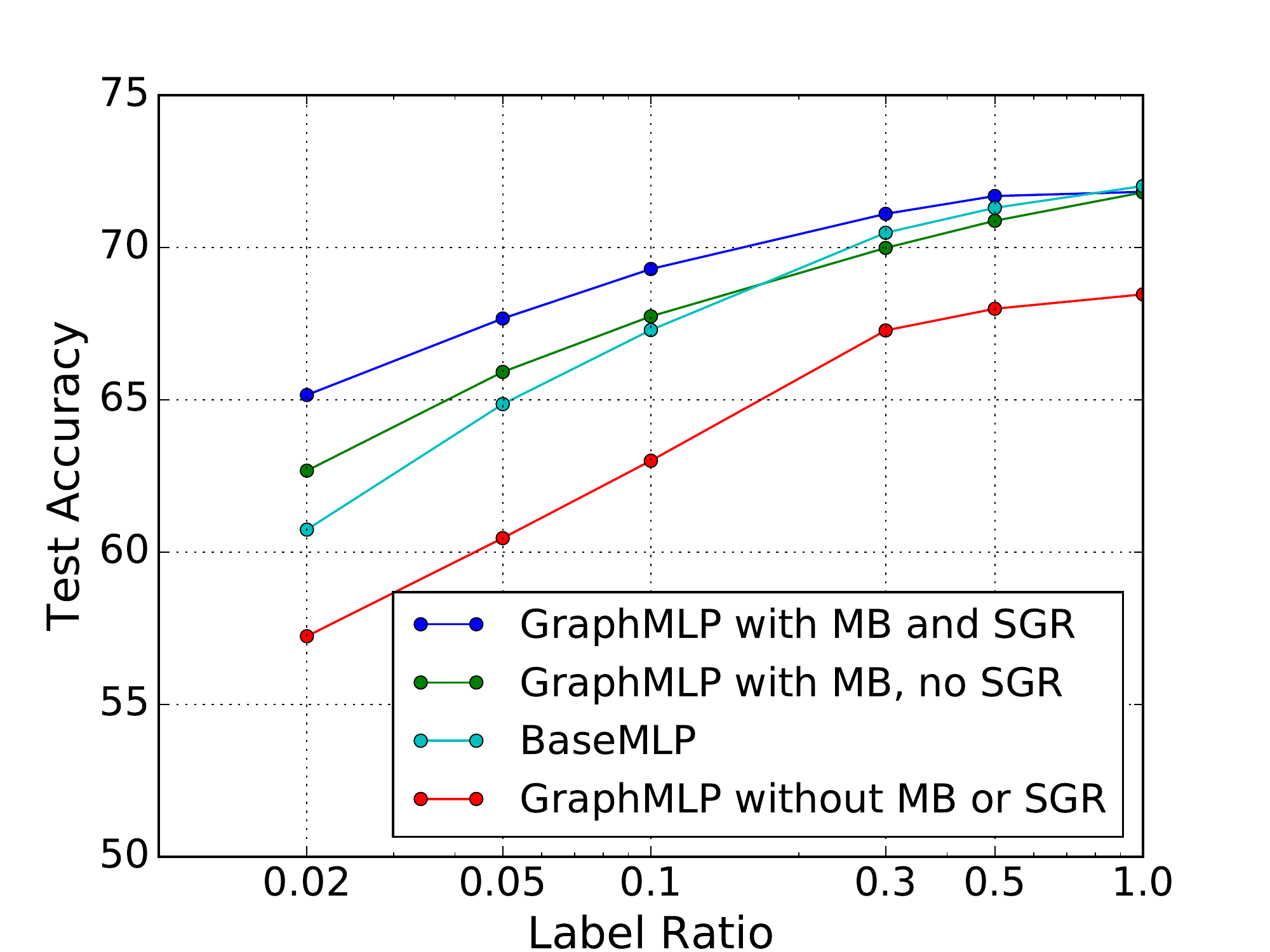}
		\caption{Effect of meta-batches and stochastic graph regularization (SGR) on the performance of a graph-regularized MLP. Also shown are results of a base-line MLP, a fully-supervised learner that only uses labeled data.}
		\label{fig:mlp_results}	
	\end{subfigure}
	\hfill
	\begin{subfigure}{0.66\columnwidth}
		\includegraphics[width=\columnwidth]{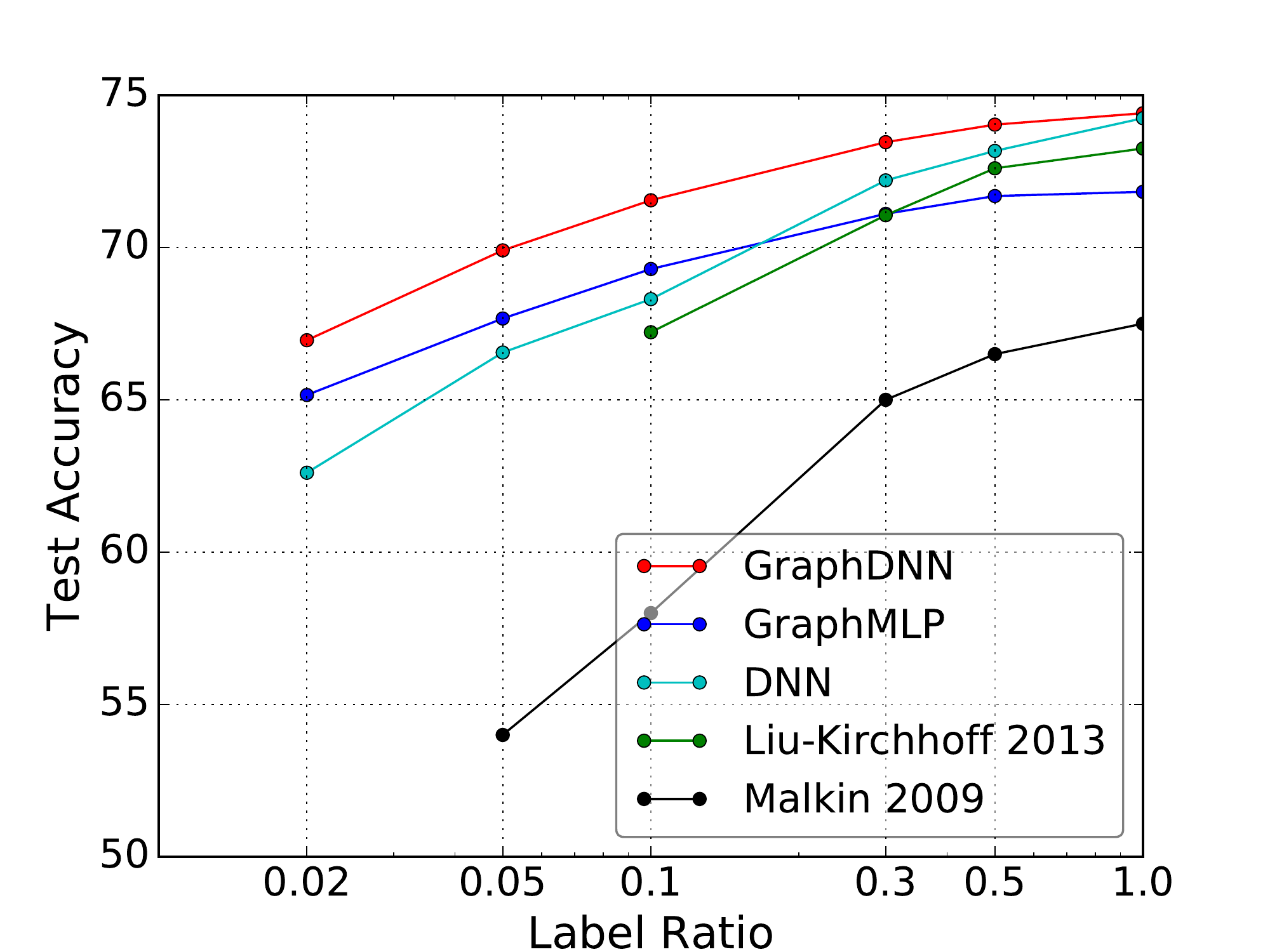}
		\caption{Results for Graph Regularized DNN vs base-line DNN. The DNN used four hidden layers, each 2000 units, with dropout. Also shown are results for similar experiments in the literature that used the same TIMIT training and test data.}
		\label{fig:dnn_results_2}
	\end{subfigure}
	\hfill
	\begin{subfigure}{0.66\columnwidth}
		\includegraphics[width=\columnwidth]{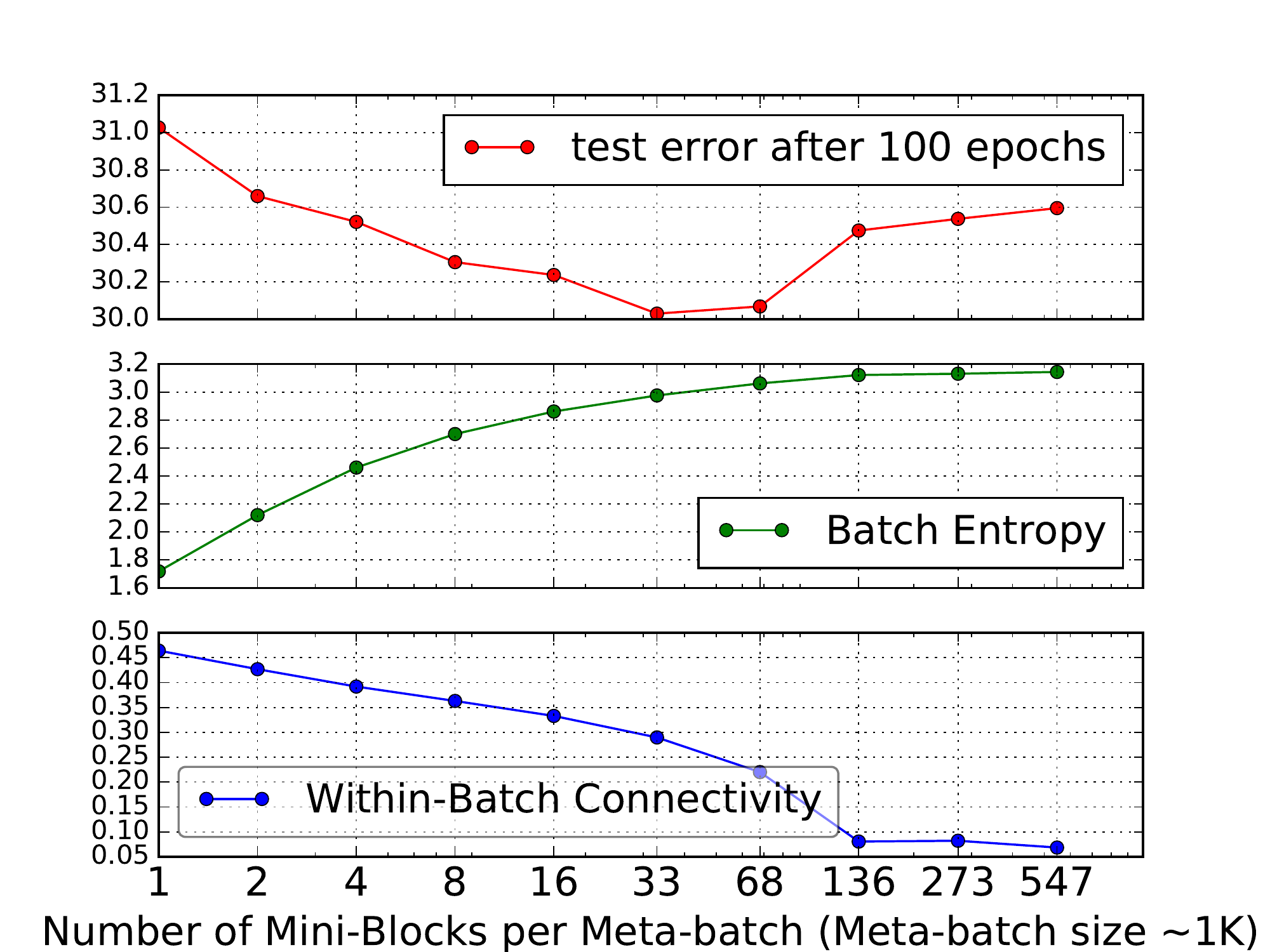}
		\caption{Trade-off between entropy and within-batch neighbor connectivity, and the performance (in terms of test error) at each of these scenarios. We used $5 \%$ of the labels, a batch-size of $\approx 1095$ and trained for  $100$ epochs. }
		\label{fig:trade_off_results}
		\end{subfigure}
\caption{}
\label{}
\end{figure*}


For all our experiments in this work 
  we use the TIMIT speech corpus~\cite{garofolo1993darpa} and just report the frame-level phone classification accuracy. Features consist of $39$-d vectors consisting of MFCC coefficients, and first and second derivatives. All data is normalized for zero mean and unit variance. We apply a sliding window of radius $4$, resulting in a  351 dimensional feature vector. The output is a distribution over $49$ classes, which is collapsed to $39$ classes during scoring. We use the 362 speaker set for training and experiment with  label ratios of $2\%,~5\%,~10\%,~30\%,~50\%$ and $100\%$ by randomly dropping labels from our training set.   Hyper-parameters were tuned using parallel grid search on a validation set.  We implemented all our models using the Theano toolkit~\cite{bergstra2010theano}.  
For the results reported here we used the AdaGrad~\cite{duchi2011adaptive} variant of gradient descent and use a hold-out  set for early stopping. For the $k$-NN graph construction, we set $k=10$ for all the experiments and use the Scikit machine learning library \cite{pedregosa2011scikit} that constructs the graphs using a fast ball-tree search. After symmetrization, affinities are computed by applying a radial basis function (RBF) kernel, such that  each entry $w_{ij}$ in the affinity matrix $W$, $w_{ij} = e^{-\frac{||x_i-x_j||}{2\sigma^2}}$. 
$\sigma$ controls the width of the kernel and determines how quickly the influence of a neighbor node decays with distance.  As in~\cite{malkin2009semi}, we tune $\sigma$ over the set $\{d_i/3\}$ where $i \in \{1,2,3,4,5\}$ and $d_i$ is the  average distance between a node and it's $i$-th nearest neighbor. For graph partitioning, we use the  METIS graph partitioning library~\cite{karypis1998multilevelk} that uses a recursive multi-way partitioning to give approximately balanced blocks.

We initially tested the benefit of the meta-batches and stochastic graph regularization on a shallow neural network -- a multi-layer perceptron (MLP) having one hidden layer of $2000$ units and a softmax output layer. These results are shown in Figure~\ref{fig:mlp_results}. A graph-regularized MLP that uses mini-batches based on purely graph partitions, and without additional out-of-batch neighbor regularization performs the worst (red curve in Figure~\ref{fig:mlp_results}); this is not surprising considering the biased gradients when using relatively homogeneous graph-based mini-batches. Using meta-batches, both with and without stochastic out-of-batch regularization (the blue and green curves respectively), noticeably improves performance, the former beating the base MLP (a supervised learner) at all scenarios except the fully-labeled case. Next, we conducted experiments on a DNN with four hidden layers, each 2000 units wide, using Rectified Linear Units~\cite{zeiler2013rectified} as the non-linear activation function, and a softmax output layer. We used dropout while training, reporting the results for the case when dropout probability is $0.2$, for which we saw the best performance.  Dropout is essentially a stochastic regularization technique and admittedly changes our objective function, but it is interesting to note that even in this setting, the graph regularization still significantly improves classification performance over the baseline DNN at the lower label ratios as shown in Figure~\ref{fig:dnn_results_2}. We also compare against the results reported in ~\cite{malkin2009semi}, which was the main reference point in this work.  In addition to the optimizations described in this paper, compared to~\cite{malkin2009semi} we can also train for significantly longer number of epochs owing to both greater computational capacity and better adaptive gradient methods which allows us to improve over the results in that paper. We also provide a comparison against a somewhat similar (although non-parametric) graph-based SSL framework reported in ~\cite{liu2013graph}.  Compared to the latter work, for the MLP case, we get better phone accuracy rates although at higher label ratios~\cite{liu2013graph} is better. This is probably due to regularizing against a prior distribution output from a first-pass classifier, which provides better priors at higher label ratios. When moving to  DNNs, however, we are able to improve performance over ~\cite{liu2013graph}. 

 Finally in Figure~\ref{fig:trade_off_results}, we illustrate empirically the trade-off between stochasticity (measured in terms of label entropy) and batch-connectivity when constructing meta-batches. For a meta-batch of a fixed size, which represents our memory budget (results shown are for batch size $\approx$ 1095, and for a labeled-to-unlabeled ratio of $0.05$), we construct meta-batches with varying entropy (diversity) by varying the size of the constituent mini-blocks. This in turn affects the neighbor connectivity (shown in the bottom plot); meta-batches constructed from larger constituent mini-blocks have lower diversity but higher neighbor density, and conversely, higher diversity and lower neighbor-connectivity  when the constituent mini-blocks are smaller.  To illustrate the result of the trade-off, we plot the test-error after $100$ epochs of training (which represents our computational budget). The test-error performance curve improves as we increase the diversity of the meta-batch, but beyond a certain point the connectivity score becomes poor enough that the benefit of additional diversity is outweighed by the lack of neighbors against which to perform graph regularization.  Characterizing this trade-off in more detail in terms  of batch diversity and neighbor connectivity is one of the focal areas of our current research.

\section{Current Research Directions and Concluding Remarks}
\label{sec:conclusion}

We presented an efficient method for graph-based SSL learning. The meta-batch construction method preserves computational efficiency and also preserves enough stochasticity in the training samples for convergence to good solutions. The stochastic graph regularization technique allows  efficient out-of-batch regularization
 and though we only report frame-level classification accuracy on the TIMIT set, this is a general, scalable  technique that can be  applied to much larger data sets, other types of classifiers and even to online learning settings. 
In addition to the graph embeddings described in this report,  we also plan to construct embeddings using a supervised pre-training phase as well as using de-noising auto-encoders~\cite{vincent2008extracting} which might provide better measures of similarity especially when labeled ratios are very small. 
We  also plan to conduct more experiments using convolutional neural networks, and experiment with much larger data sets.

  \eightpt
  \bibliographystyle{IEEEtran}
  
  \bibliography{deep_ssl}

\end{document}